\documentclass{article}

\usepackage{arxiv}
\usepackage{graphicx}
\usepackage[hyphens]{url}  % DO NOT CHANGE THIS
\usepackage[numbers]{natbib}  % DO NOT CHANGE THIS AND DO NOT ADD ANY OPTIONS TO IT
\usepackage{caption} % DO NOT CHANGE THIS AND DO NOT ADD ANY OPTIONS TO IT
\usepackage{booktabs}
\usepackage{amsmath}
\usepackage{amsfonts}
\usepackage{stmaryrd}
\usepackage{hyperref}
\usepackage[capitalize,noabbrev]{cleveref}
\usepackage[disable]{todonotes}
\usepackage{newfloat}
\usepackage{multirow}

\newcommand{\orcidID}[1]{\href{https://orcid.org/#1}{\includegraphics[scale=0.06]{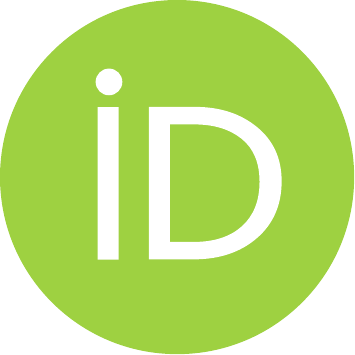}\hspace{1mm}}}

\renewcommand{\headeright}{An Extended Preprint}
\renewcommand{\undertitle}{An Extended Preprint}

\definecolor{DarkBlue}{rgb}{ 0 , 0 , 0.7}

\begin{document}
\title{Tackling Morphological Analogies Using Deep Learning -- Extended Version}
\author{\orcidID{0000-0002-4132-1068}Safa Alsaidi\\Université de Lorraine, CNRS, LORIA\\ F-54000, France \And
\orcidID{0000-0001-6773-9983}Amandine Decker\\Université de Lorraine, CNRS, LORIA\\ F-54000, France \And
\orcidID{0000-0003-2315-7732}Esteban Marquer\\Université de Lorraine, CNRS, LORIA\\ F-54000, France\\
\texttt{esteban.marquer@loria.fr} \And
\orcidID{0000-0003-4586-9511}Pierre-Alexandre Murena\\
HIIT, Aalto University\\ Helsinki, Finland\\
\texttt{pierre-alexandre.murena@aalto.fi} \And
\orcidID{0000-0003-2316-7623}Miguel Couceiro\\Université de Lorraine, CNRS, LORIA\\ F-54000, France\\
\texttt{miguel.couceiro@loria.fr}
}
% First names are abbreviated in the running head.
% If there are more than two authors, 'et al.' is used.
%
%
\maketitle              % typeset the header of the contribution
\begin{abstract}
Analogical proportions are statements of the form ``$A$ is to $B$ as $C$ is to $D$''.
They constitute an inference tool that provides a logical framework to address learning, transfer, and explainability concerns and that finds useful applications in artificial intelligence and natural language processing.
In this paper, we address two problems, namely, analogy detection and resolution in morphology.
Multiple symbolic approaches tackle the problem of analogies in morphology and achieve competitive performance.
We show that it is possible to use a data-driven strategy to outperform those models.
We propose an approach using deep learning to detect and solve morphological analogies.
It encodes structural properties of analogical proportions and relies on a specifically designed embedding model capturing morphological characteristics of words.
We demonstrate our model's competitive performance on analogy detection and resolution over multiple languages.
We provide an empirical study to analyze the impact of balancing training data and evaluate the robustness of our approach to input perturbation.
\keywords{Analogy  solving \and Analogy  detection \and Neural networks \and Classification \and Regression \and Morphological word embeddings.}
\end{abstract}
\section{Introduction}
% embedding (W2V, glove) ~> semantic analogy ~> issues ~> Lim et al
% Even if semantic is ``solved'', morpho has improvement room 
% until now, due to the nature of the pb, exclusively symbolic (Murena et al ...), no data-driven
% Use similar model as Lim et al and achieve competitive performance / outperform state of the art

There is a long tradition of works on word representations, called \textit{word embeddings}.
%Such methods go from early methods based on word co-occurrence like latent semantic analysis~\cite{lsa:1988:dumais} to recent works involving more complex models shuch as Bert Multilingual~\cite{bert-multilingual:2019:delvin}), and including key works like GloVe~\cite{glove:2014:pennington}, Word2Vec~\cite{efficient-representation-w2v:2013:mikolov} and fasttext~\cite{fasttext:2016:bojanwoski}.
Famous examples of such methods include GloVe~\cite{glove:2014:pennington}, \textit{word2vec}~\cite{efficient-representation-w2v:2013:mikolov} and \textit{fasttext}~\cite{fasttext:2016:bojanwoski}.
It is usually assumed that the embedding space defined by those word embedding models represents the semantics of the words, and that simple arithmetic operations account for simple relations between concepts, like $\vec{cat}-\vec{kitten}+\vec{dog}\approx\vec{puppy}$~\cite{efficient-representation-w2v:2013:mikolov,embedding-analogies:2016:drozd,eval-vector-analogy:2017:chen}.
The above relation is very close to \textit{analogical proportions}, statements of the form ``$A$ is to $B$ as $C$ is to $D$'', usually linked with \textit{analogical equations} of the form ``what is to $C$ as $B$ is to $A$''.
However, the aforementioned embeddings have not been trained explicitly to find analogies, and the fact that such relations can be observed is an emerging effect of the semantic representation.
In addition, it has also been found that such simple approaches are not enough to reach human performance in all cases~\cite{eval-vector-analogy:2017:chen}.

Based on a formalism describing analogical proportions~\cite{trends-analogical-reasoning:2014:prade,analogical-dissimilarity:2008:miclet}, Lim~\textit{et~al.}~\cite{analogies-ml-perspective:2019:lim} proposed a deep learning model to tackle such analogies using \emph{semantic} embeddings. Unlike for the previous works, the architecture is adapted to the characteristics of analogies and the model is trained using a dataset of analogies.  
%
%Even though the problem of analogies on semantic embeddings appears to be solved, analogies on character strings and in particular in morphology has margin for improvement.
Although this approach seems to solve the problem of analogies on semantic embeddings, analogies on character strings (in particular in morphology) has still margin for improvement.
Indeed, while recent approaches using symbolic methods \cite{analogy-alea:2009:langlais,tools-analogy:2018:fam-lepage,minimal-complexity:2020:murena} achieve competitive performance on detection of analogies and solving of analogical equations, they are inherently restricted to a subset of morphology.
% put some use cases for morphological analogies, to show that it is useful

%Similarly to the method of Lim~\textit{et~al.}~\cite{analogies-ml-perspective:2019:lim}, we propose neural models that learn, from analogies extracted from data, to detect and solve analogies in the morphology domain. To do so, the two crucial steps are the learning of an embedding particularly adapted to words seen as character strings, and the definition of a network adapted to the formal properties of analogy.
%%%% We propose a data-driven model similar to the one of \citet{analogies-ml-perspective:2019:lim} to tackle morphological analogies, and achieve competitive performance on both analogy detection and solving.
% In \cref{sec:formal-analogy} we detail the setting of analogy in morphology and detail key approaches.
% The models we propose to identify and solve analogies, as well as the procedures we use for training and evaluating them, are described in \cref{sec:model}.
% We detail some results on the data introduced in \cref{sec:data}\todo{check} on the detection and solving tasks in \cref{sec:clf-expe} and \cref{sec:reg-expe} respectively, and present in \cref{sec:dropout} experimental results of the impact on our models' performance of applying noise on the embeddings.

%(Esteban) we need to make the particularities of the model clear
%- embedding
%- analysis
%- morphology (regression, see previous work for clf)
We propose models that learn, from analogies extracted from data, to detect and solve analogies in the morphology domain. To do so, the two crucial steps are the learning of an embedding particularly adapted to words seen as character strings, and the definition of a network adapted to the formal properties of analogy. Lim~\textit{et~al.}~\cite{analogies-ml-perspective:2019:lim} have shown the performance of a similar deep learning framework on pre-trained semantic word embeddings for analogies on word semantics, and we argue that the framework itself has the potential to be applied to analogies in a wide range of domains and applications.
%As discussed in~\cite{transfer:2021:alsaidi}, we could further extend the framework.

In \cref{sec:formal-analogy} we detail the setting of analogy in morphology and detail key approaches.
The models we propose to identify and solve analogies and the intuitions leading to the choice of the architecture are described in \cref{sec:model}, along with the procedures we use for training and evaluating them.
We detail some results on the multilingual data introduced in \cref{sec:data} on the detection and solving tasks in \cref{sec:clf-expe} and \cref{sec:reg-expe} respectively, and present in \cref{sec:dropout} experimental results of the impact on our models' performance of applying noise on the embeddings.

The main contributions of this paper are as follows.
\begin{itemize}
    \item We propose a word embedding model for analogy on morphological features.
    \item We display the performance of our deep learning models, which outperform state of the art symbolic approaches to detect and solve analogies on morphological features.
%    \item We evaluate the performance of our models on 11 languages and explore their sensibility to perturbations in the embedding space.
    \item We demonstrate the importance of a correctly chosen embedding space for analogy manipulation and display the potential of the approach as a general framework for detecting and solving analogies.
\end{itemize}
%%%%%%%%%%%%%%%%%%%%%%%%%%%%%%%%%%%%%%%%%

\section{The Problem of Morphological Analogy}\label{sec:formal-analogy}
%same as DSAA maybe more compressed:
%part A OR AIMLAI 2.1 + more examples 
%skip part C

An analogical proportion is defined as a 4-ary relation written $A:B::C:D$ and which reads ``$A$ is to $B$ as $C$ is to~$D$". In this paper, we focus more specifically on the case of morphological analogies, \textit{i.e.}, on analogies involving character strings, where the transformations between the objects correspond to morphological transformations of words (\textit{e.g.}, conjugation or declension). In our setting, $A$, $B$, $C$, and $D$ are words.

\paragraph{Proportional analogy.}
 The importance of analogy in morphology has been long known~\cite{linguistique-generale:1916:saussure}, and it has been mathematically formalized following the early works of Lepage~\cite{lepage1996saussurian,analogy-commutation-linguistic-fr:2003:lepage}. These works paved the way toward the current definition of \emph{proportional analogy}~\cite{analogical-dissimilarity:2008:miclet,trends-analogical-reasoning:2014:prade} as a 4-ary relation for which the following postulates hold true for all $A$, $B$, $C$, and $D$:
\begin{itemize}
    \item $A : B :: A : B$ (reflexivity);
    \item $A : B :: C : D \rightarrow C : D :: A : B$ (symmetry);
    \item $A : B :: C : D \rightarrow A : C :: B : D$ (central permutation).
\end{itemize}
It can be easily verified that other equivalent forms can be derived by applying the symmetry and central permutation: $D:B::C:A$, $C:A::D:B$, $B:A::D:C$, $D:C::B:A$ and $B:D::A:C$.

We identify two main trends in the research on analogies: \emph{analogy detection}, which corresponds to deciding whether a quadruple $\langle A, B, C, D \rangle$ is a valid analogy, and \emph{analogy  solving}, which corresponds to finding the values of $x$ such that $A : B :: C : x$ is a valid analogy. We note here that these tasks can be interpreted as respectively classification and regression tasks.

\paragraph{Analogy detection.} The analogy detection task corresponds to learning a classifier classifying quadruples into valid or invalid analogies. In particular, in a context of semantic analogies, Bayoudh~\textit{et~al.}~\cite{bayoudh2012evaluation} proposed to use Kolmogorov complexity as a distance measure between words in order to define a conceptually significant analogical proportion and to classify valid or invalid analogies. A completely functional form was implemented by Lim~\textit{et~al.}~\cite{analogies-ml-perspective:2019:lim}, which propose to learn the classifier directly in the form of a neural network. 
Analogy detection has been used in particular in a context of analogical grids~\cite{tools-analogy:2018:fam-lepage}, \textit{i.e.}, matrices of transformations of various words, similar to paradigm tables in linguistic~\cite{fam2016morphological}.

\paragraph{Analogy solving.} The analogy solving tasks corresponds to inferring the fourth term that makes an analogy valid. This task is by far the most investigated in the analogy literature and is solved by a variety of approaches.

In the context of solving morphological analogies, the major trend follows the seminal works of Lepage and Ando~\cite{lepage1996saussurian} and exploits the axioms of proportional analogy introduced above. For instance, Lepage~\cite{sigmorphon:2017:lepage} proposes to apply the axioms of proportional analogy to multiple characteristics of the words, such as their length, the occurrence of letters and of patterns. 
Following the results of Yvon~\cite{finite-state-trancducers:2003:yvon} about the closed forms the solution of a proportional analogy can have, the \emph{alea} algorithm~\cite{analogy-alea:2009:langlais} proposes a Monte-Carlo estimation of the solutions of an analogy, by sampling among multiple sub-transformations. 
Recently, a more empirical approach was proposed by~Murena~\textit{et~al.}~\cite{minimal-complexity:2020:murena}, which does not rely on the axioms of proportional analogy. Their method consists in finding the simplest among all possible transformations from $A$ to $B$ which could apply on $C$ as well. 

% Finally, we would like to insist on the fact that a very particular case of proportional analogy in vector spaces is the \emph{parallelogram rule}, which states that four vectors $A, B, C$ and $D$ are in analogical proportion if $D - C = B - A$. This relation has been used since the first works on analogy~\cite{rumelhart1973model} and is a key element of the methodology employed by the recent neural-based approaches~\cite{efficient-representation-w2v:2013:mikolov,embedding-analogies:2016:drozd}, where it is applied in some learned embedding space.  
%\todo{Add the pb itself: incorrect perf wrt. humans, ~> Lim bring notion of solution (or maybe not, as it is a reapeat of what is said in intro?)}
%\todo{This paragraph is not suited for "analogy solving" as it corresponds to both solving and detection. Transfer to next paragraph?}

% \paragraph{Our solution.} Similarly to the method of~\citet{analogies-ml-perspective:2019:lim}, our method aims to learn a functional form for both the analogy detection and analogy solving tasks. \textbf{TODO}
%\todo{Do we want to give more details here? (E) No, but we can use this as a transition}

\paragraph{Analogy in embedding spaces.}
Finally, we would like to mention that a very particular case of proportional analogy in vector spaces is the \emph{parallelogram rule}, which states that four vectors $A$, $B$, $C$, and $D$ are in analogical proportion if $D - C = B - A$. This relation has been used since the first works on analogy~\cite{rumelhart1973model} and is a key element of the methodology employed by the recent neural-based approaches~\cite{efficient-representation-w2v:2013:mikolov,embedding-analogies:2016:drozd}, where it is applied in some learned embedding space. 
However, as mentioned in introduction, the results produced by this method differ from human performance~\cite{eval-vector-analogy:2017:chen}, and frameworks based on datasets of analogies like~\cite{analogies-ml-perspective:2019:lim} appear to bridge this gap in performance.
Similarly to the method of Lim~\textit{et~al.}~\cite{analogies-ml-perspective:2019:lim}, our method aims to learn a functional form for both the analogy  detection and solving tasks in morphology. 
In particular, this form enables to cover the actual relations of analogies, unlike the \textit{ad hoc} methods (like parallelogram rule) usually used in embedding spaces. 
%Instead of using an \textit{ad hoc} formula/method (like parallelogram) that happens to match some analogies, our model learns the actual relations of analogies

\section{Proposed Approach}\label{sec:model}
\begin{figure*}[htb]
    \centering
    \includegraphics[width=\textwidth]{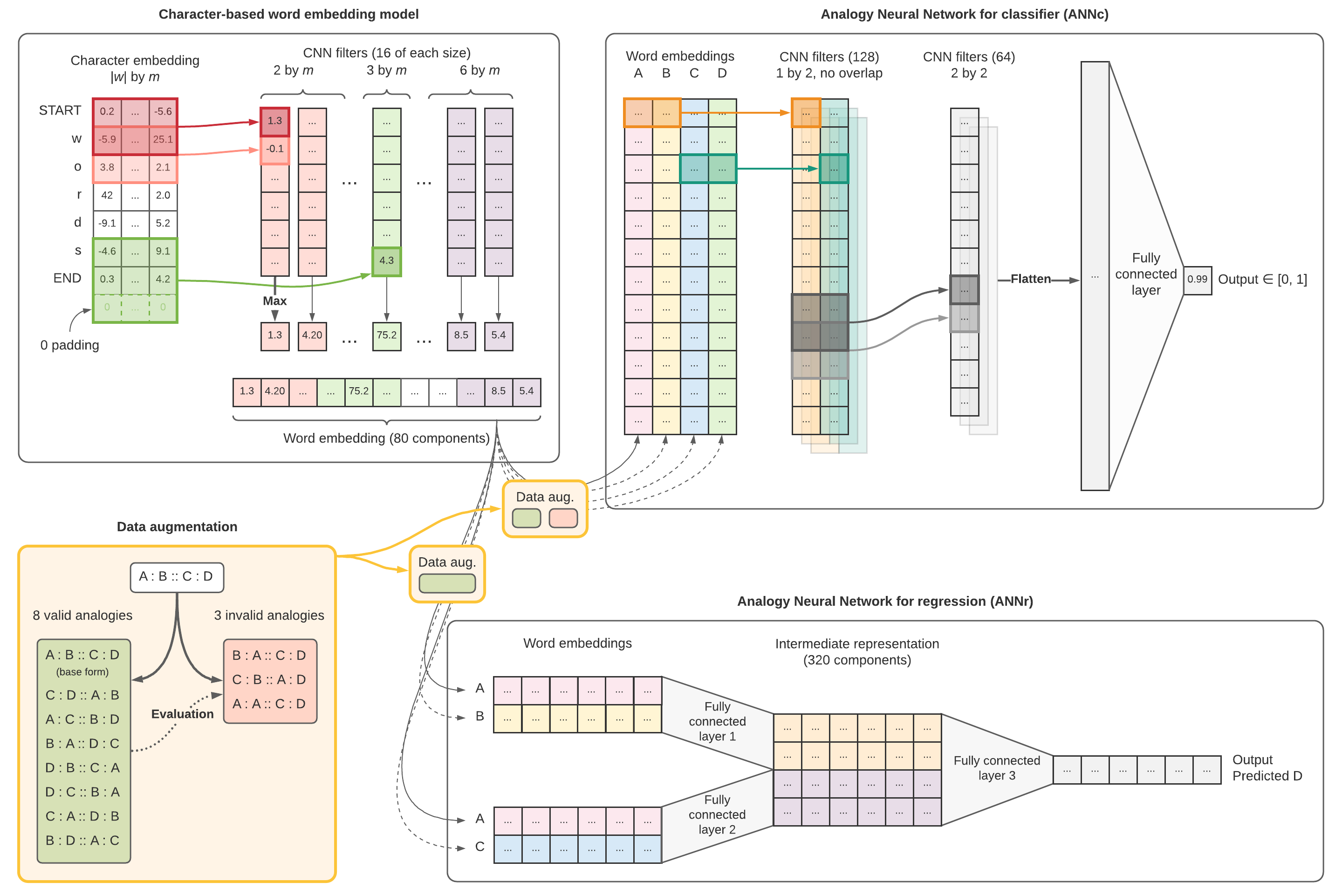}
    
    \caption{Morphological embedding model, data augmentation, analogy classification and regression models. Note that invalid analogies are used only for the classifier.}
    \label{fig:model}
\end{figure*}

This section describes the proposed approach, illustrated in \cref{fig:model}.
There are 3 components in our models: a classification model and a regression model described in \cref{ssec:clf-reg}, both taking as input embeddings computed by our morphological embedding model presented in \cref{ssec:emb}. 
Then, we detail our training and evaluation procedures in \cref{sec:training} and give additional information about our model hyperparameters in \cref{ssec:hyperparams}.%\todo{respecify pb}

%------------------------------------------------
\subsection{Classification and Regression Models}\label{ssec:clf-reg}
%using model from Lim et al.... (a few lines only here), describe and explain it a bit
The models described in this subsection follow the structure of the ones proposed in~\cite{analogies-ml-perspective:2019:lim} and are schematized in \cref{fig:model}.

\paragraph{Classification.} %As an analogy $A:B::C:D$ is valid if $A$ and $B$ differ in the same way as $C$ and $D$, the link between $A$ and $B$ and the one between $C$ and $D$ are essential. Thus, a possible view of the classification task is to approximate $g$ such that $g(f_1(A,B), f_2(C,D))$ determining the validity of $A:B::C:D$, where $f_1$ and $f_2$ describe respectively the link between $A$ and $B$, and $C$ and $D$.%\todo{If only there was a way to put this !?\#!\$ $D$ on the previous line...}
%
%The classification neural network proposed follows this idea.
%Our classification neural network follows the idea that a possible view of the classification task is to approximate $g$ such that $g(f(A,B), f(C,D))$ determining the validity of $A:B::C:D$, with $f$ describing the link between two elements.
Our classification neural network follows the idea that an analogy $A:B::C:D$ is valid if $A$ and $B$ differ in the same way as $C$ and $D$.
A first Convolutional Neural Network (CNN) layer with 128 filters of $1$~{by}~$2$ is applied on the embeddings, such that it spans over $A$ and $B$ on one side and $C$ and $D$ on the other side, for each component of the embedding, without overlap. This first layer measures how $A$ and $B$ differ for each component, and similarly for $C$ and $D$. %This layer corresponds to $f$. 
Then, a second CNN layer with 64 filters of $2$~{by}~$2$ is applied on the resulting matrix, followed by a fully connected layer producing a single output. 
% These two layers correspond to $g$, and the output is the classification result normalized between 0 (\textit{i.e.} invalid) and 1 (\textit{i.e.} valid) thanks to a sigmoid activation.
This second CNN layer allows to check if the difference between $A$ and $B$ is the same as the one between $C$ and $D$, and the last layer aggregates the information.
The output is the classification result normalized between 0 (\textit{i.e.}, invalid) and 1 (\textit{i.e.}, valid) by a sigmoid activation.
All layers except the output layer use Regularized Linear Unit (ReLU) as activation function. We call this model Analogy Neural Network for classification (ANNc).

\paragraph{Regression.}
A simplified view of solving $A:B::C:x$ is to find how $B$ differs from $A$ and generate $D$ to differ in the same way from $C$.
\textit{Central permutation} (see \cref{sec:formal-analogy}) allows us to apply the same operations on $A:C::B:x$, to obtain $D$ from $B$ using the difference between $A$ and $C$. % with $B$ to obtain $D$.
%A simplified view of solving $A:B::C:x$ is to find how $B$ differs from $A$ and generate $D$ to differ in the same way from $C$, and \textit{central permutation} (see \cref{sec:formal-analogy}) allows us to perform in the same way with $A$ and $C$, and $B$ and $D$.\todo{reformulate}
Our regression model first determines the relation between $A$ and $B$ on one side and the one between $A$ and $C$ on the other side, by applying two separate fully connected layers with ReLU activation on the respective embeddings ($A$ and $B$ for one layer, and $A$ and $C$ for the other). Those layers also extract the necessary information from $B$ and $C$ respectively, and are able to take $A$ into account in that process.
Finally, a third fully connected layer without activation generates the embedding of the predicted $D$ from all the extracted information.
We call this model Analogy Neural Network for regression (ANNr).

\subsection{Embedding Model}\label{ssec:emb}
%tests with Glove \& W2V \& fasttext on classification German: very bad performance, ours >>> fasttext > Glove > W2V
%fasttext is already based on sub-words
\paragraph{Pre-trained embeddings.}
An important component of deep learning models applied on textual data is what is called the embedding model, which produces real-valued vector representation of the textual data, called \textit{embedding}.
Many applications rely on embedding models that have been extensively pre-trained on large amounts of data, as it avoids the cost of fully training one oneself.
It is commonly accepted that the quality of this representation, which corresponds to the amount and nature of the encoded information as well as the properties of the embedding space, is a key factor to achieve higher performance.

We trained the classification model on a subset of the German data of Sigmorphon2016~\cite{sigmorphon:2016:cotterell} (\cref{sec:data}) with pre-trained embeddings such as GloVe~\cite{glove:2014:pennington}, \textit{word2vec}~\cite{efficient-representation-w2v:2013:mikolov}, and \textit{fasttext}~\cite{fasttext:2016:bojanwoski}, which are known to encode semantic information from distributional properties, our ANNc performed poorly.
It is noteworthy that \textit{fasttext}, which uses sub-words instead of words, produced better results than GloVe or \textit{word2vec}.
Such general-purpose embedding models, oriented towards the semantics of words, seem ill-suited to manipulate morphological analogies.
We thus propose a word embedding model more suited to handle morphological tasks in the next paragraph.
%Additionally, all the languages we might manipulate do not have pre-trained embedding models available.

\paragraph{Morphological embeddings.}
To handle morphological information, we designed an embedding model inspired from the work of Vania~\cite{morpho-thesis:2020:vania}.
As represented in \cref{fig:model}, the input of our embedding model is the characters of a word, which are embedded into vectors of size $m$ using a learned embedding matrix.

Multiple CNN filters go over the character embeddings, spanning over the full embeddings of $2$ to $6$ characters, resulting in filter sizes between $2$ by $m$ and $6$ by $m$.
We use 16 filters of each size, and the convolution is performed by sliding the filters over the character dimension.
To allow the filters to detect the beginning and end of the word, %two special characters are added at the extremities of the word.
special characters are added at both ends of the word.
Additionally, to allow smaller words to accommodate larger filters, $0$ padding is used.
For each filter, the model computes the maximum of the output, which will serve as a component of the word embedding.

The intent of this last operation is to keep only salient patterns detected by each of the filters, and force each CNN filter to specialize in identifying a specific pattern of characters.
As we use character embeddings, the model can capture not only patterns such as ``-ing'' but also ``vowel-vowel-consonant'', as some features of the characters can be extracted.
These character patterns correspond to what is called \textit{morphemes}, minimal units of morphology.

%-----------------------------------
\subsection{Training and Evaluation}
\label{sec:training}
%same as in AIMLAI or DSAA

\paragraph{Data Augmentation.}
The properties of formal analogy described in \cref{sec:formal-analogy} allow us to generate multiple analogies from each base example in our dataset in what we call \textit{data augmentation}.
This process allows us to augment the amount of data available, but also to make the models learn to fit the formal constraints of analogy.
Given a valid analogy $A:B::C:D$, we  generate 7 more valid analogies, namely, $A:C::B:D$, $D:B::C:A$, $C:A::D:B$, $C:D::A:B$, $B:A::D:C$, $D:C::B:A$, $B:D::A:C$. These 8 forms are used for the classification and the regression task.
Additionally, we can generate 3 analogical forms per valid analogy which are considered invalid analogies as they cannot be deduced from the base form $A:B::C:D$ without contradicting at least one of the postulates mentioned in \cref{sec:formal-analogy}: $B:A::C:D$, $C:B::A:D$, and $A:A::C:D$. Those 3 forms are used to generate negative examples for the classification task, and can be applied on any of the 8 equivalent forms.
In total, for each analogy in the dataset, we can generate $8$ valid and $8\times3=24$ invalid analogies for the classification task.

For evaluating the classifier, we use the full set of 24 invalid analogies, to have as many cases as possible.
However, for training the classifier, we determined that balancing the data by using 8 invalid analogies sampled from the 24 available was better than using either the full 24 invalid analogies (strong imbalance, 3 times more invalid than valid) or only the 3 invalid analogies obtained from the base form (weaker imbalance, 8 valid for 3 invalid). 
Detailed experimental results and significance tests are available in the appendix\todo{add to appendix}.
%The detail of the experimental results and significance tests are available in the appendix\todo{add to appendix}.

%Through several experiments we determined that training using 8 invalid analogies sampled from the 24 available was better than using either the full 24 invalid analogies (strong imbalance, 3 times more invalid than valid analogies) or only the 3 invalid analogies obtained from the base form (imbalance, 8 valid for 3 invalid). The detail of the experimental results and significance tests are available in appendix\todo{add to appendix}.

In practice, data augmentation is applied after the embedding model for both the classification and the regression model, as shown in \cref{fig:model}.

% For evaluation we use for each 
% eval on 8by24 to have all the possible forms
% 8by8 better than 8by3 and 8by24 (see appendix for detailed result)
% - significative difference between 8x24 and the other two for positive (i.e 8x8 and 8x3 are significantly better than 8x24 on positive)
% - significative difference between 8x3 and the other two for negative (i.e 8x8 and 8x24 are significantly better than 8x3 on negative)
% - almost significative difference between 8x8 and 8x24 on negative, which paired with the results may indicate that for some languages, using all 24 is slightly better (which makes sense), yet overall sampling 8 is enough to reach the same performance;
% - non-significative difference (> 0.1) between 8x8 and 8x3 on positive, indicating similar performances.
% Overall, the statistical test does enforce our conclusion that 8x8 is better as it reaches the performance of 8x3 on positive while being very close to the performance of 8x24 on negative.

% For regression, only the 8 positive

\paragraph{Optimization criterion.}
For classification, we use the Binary Cross-Entropy (BCE) loss (see \cref{equ:bce} below) to minimize the distance between the expected label $l\in \{0,1\}$ and the predicted label $\hat{l}\in [0,1]$.
\begin{equation}
    BCE(l, \hat{l}) = l \ \log(\hat{l}) + (1 - l) \ \log(1 - \hat{l}) \label{equ:bce}
\end{equation} 

For regression, we minimize the ratio between: (\textit{i}) the distance between the actual embedding $e_D$ and the predicted embedding $\widehat{e_D}$ and (\textit{ii}) the average distance between the embeddings $e_A$, $e_B$, $e_C$, and $e_D$.
We minimize this ratio instead of the distance between $e_D$ and $\widehat{e_D}$ to avoid a caveat encountered during our early experiments.
Indeed, the embedding model is learned together with the regression model, and grouping all the embeddings in a small area of the embedding space also minimizes the distance between $e_D$ and $\widehat{e_D}$. The ratio we use as a criterion ensures that the distance between $e_D$ and $\widehat{e_D}$ is minimized relatively to the distance between the other elements.
The formula of our ratio:
\begin{equation}
\mathcal{L}_{reg}(e_A,e_B,e_C,e_D, \widehat{e_D}) = 
\frac{6 \times d_{D\hat{D}}}{1 + d_{AB} + d_{AC} + d_{AD} + d_{BC} + d_{BD} +  d_{CD}} \label{equ:relative-mse}
\end{equation}
uses the Mean Square Error (MSE) loss as a measure of the $L^2$ distance (see \cref{equ:mse}, where $n$ is the number of dimensions of the embeddings). In \cref{equ:relative-mse}, the MSE between $e_D$ and $\widehat{e_D}$ is written $d_{D\hat{D}}$, with a similar writing for other MSE.
\begin{equation}
MSE(e_D, \widehat{e_D}) = d_{D\hat{D}} = \frac{1}{n}\sum_{i\in\llbracket 1,n\rrbracket}(e_{D,i} - \widehat{e_{D,i}})^2 \label{equ:mse}
\end{equation} 

\paragraph{Inference and evaluation.}
To evaluate the ANNc and ANNr, we use classification and regression accuracy respectively.
For the regression model, the output is not necessarily the exact embedding of a word but a real-valued vector in the embedding space. As the model is not equipped with a decoder for the embedding space, we do not have direct access to the word corresponding to a generated embedding.
Instead, we take the closest embedding among the embeddings of the whole vocabulary; the corresponding word will be considered as the prediction of the model.
To select the closest embedding, we use Euclidean distance.
We experimented with both Euclidean and cosine distance, the former giving slightly better results in most cases, even though the difference is not very significant.

%--------------------------------------
\subsection{Hyperparameters and Tuning}\label{ssec:hyperparams}
We use a character embedding size ($m$ in \cref{fig:model}) of $64$ for all our models except when Japanese is involved in the training, in which case we use $512$.
For all languages, we use 16 filters of each of the sizes between 2 and 6, resulting in embeddings of 80 dimensions.
Training was performed using Adam for 20 epochs~\cite{adam:2014:kingma}.
We used the recommended learning rate of $10^{-3}$ for classification, and, for regression, a learning rate of $10^{-4}$ which produced better results experimentally.

Each model is associated with an embedding model and is trained on a single language of the dataset described in next section\footnote{Experiments on transferring models between languages without retraining, as well as using a single embedding for all languages, were performed and are detailed in a recently accepted paper by Alsaidi~\textit{et~al.}~\cite{transfer:2021:alsaidi}.}.\todo{add to appendix}
%We consider regression more difficult than classification in our setting, as the objective for 
The embedding model for regression is initialized with the parameters of the embedding model of the classification model of the corresponding language to increase the convergence speed of the regression model. When training, the parameters of this embedding model are frozen for the first 10 epochs. This allows the regression model to first learn to use the existing embedding space before adapting the latter to fit the task better.

%%%%%%%%%%%%%%
\section{Data}\label{sec:data}

% data source (2 datasets)
% analogy extraction
% used amount, train & test split of jap (see code appendix for split snippet with random seed 42)
% note: our differs from Lepage's and Murena's version, cf appendix for overlap stats

%We worked with 11 languages from two datasets in our experiments: Spanish, German, Finnish, Russian, Turkish, Georgian, Navajo, Arabic (Romanized), Hungarian and Maltese from Sigmorphon2016~\cite{sigmorphon:2016:cotterell}, and Japanese from the Japanese Bigger Analogy Test Set~\cite{jap-data:2018:karpinska}.
For our experiments, we extracted analogies in 11 languages from two datasets: Spanish, German, Finnish, Russian, Turkish, Georgian, Navajo, Arabic (Romanized), Hungarian and Maltese from Sigmorphon2016~\cite{sigmorphon:2016:cotterell}, and Japanese from the Japanese Bigger Analogy Test Set~\cite{jap-data:2018:karpinska}.
Of the 3 subtasks in which the  Sigmorphon2016~\cite{sigmorphon:2016:cotterell} dataset is separated, namely, inflection, reinflection, and unlabeled reinflection, we use the data of the inflection subtask.
The extracted analogies were also used in our previous works on morphological analogies~\cite{transfer:2021:alsaidi,alsaidi:hal-03313556}.

The original data of each language is composed of pairs of words $\langle A,B\rangle$ (ex: $\langle$``do''$,$``doing''$\rangle$) with $B$ the morphological transformation of $A$ to obtain a set of morphological features $F$ (ex: present participle).
For example in our Finnish data, we have $A=$``lenkkitossut'', $B=$``lenkkitossuilla'', and $F=$``pos=N,case=ON+ESS,num=PL'' (the transformation corresponds to the nominative to essive cases of a noun for the plural).
For any two pairs of words $\langle A, B\rangle,\langle A', B'\rangle$ that have the same set of features ($F=F'$), we consider $A:B::A':B'$ an analogical proportion. Note that for each two pairs, we only generate one analogy, \textit{i.e.}, if we generate $A:B::A':B'$ we do not generate $A':B'::A:B$ as it will be generated by the data augmentation process. Also, analogies of the form $A:B::A:B$ will be generated as the set of features is the same ($F=F$).

For training and evaluation we use the original split of the Sigmorphon2016~\cite{sigmorphon:2016:cotterell} dataset, from which we extract the analogies independently.
For the Japanese Bigger Analogy Test Set~\cite{jap-data:2018:karpinska} however, no such split is available so we performed a random split (using a fixed random seed) in a training set of 70\% of the extracted analogies, the remaining 30\% serving as the test set.
The number of analogies available is described in \cref{tab:num-analogy}. In practice, to maintain reasonable training and evaluation times, we use up to 50000 analogies of each set, randomly selected when loading the data.%\todo{reformulate}
\begin{table}[htb]
    \centering
    \caption{Number of analogies generated for each language.}
    \label{tab:num-analogy}
    \begin{tabular}{lrr}\toprule
            \textbf{Language}   & \textbf{Train} &  \textbf{Evaluation} \\
            \midrule
            %\multicolumn{3}{c}{\textit{Sigmorphon2016~\cite{sigmorphon:2016:cotterell}}}\\
            \multicolumn{3}{c}{\textit{Sigmorphon2016}}\\
            %\multicolumn{3}{c}{\textit{\cite{sigmorphon:2016:cotterell}}}\\
            Arabic             (Ar) & 373240 & 555312 \\
            Finnish            (Fi) & 1342639 &  4691453 \\
            Georgian           (Ka) & 3553763 &  8368323 \\
            German             (De) & 994740 &  1480256 \\
            Hungarian          (Hu) & 3280891 &  66195 \\
            Maltese            (Mt) & 104883 & 3707 \\
            Navajo             (Nv) & 502637 &  4843 \\
            Russian            (Ru) & 1965533 &  6421514 \\
            Spanish            (Es) & 1425838 &  4794504 \\
            Turkish            (Tr) & 606873 &  11360 \\
            \midrule
            \multicolumn{3}{c}{\textit{Japanese Bigger Analogy Test Set}}\\
            %\multicolumn{3}{c}{\textit{\cite{jap-data:2018:karpinska}}}\\
            Japanese           (Jp) & 18487 & 7923 \\
            \bottomrule
        \end{tabular}
\end{table}

It is noteworthy that we use the data of the original Sigmorphon2016 dataset, which produces a different set of analogies than the one used by \cite{tools-analogy:2018:fam-lepage,minimal-complexity:2020:murena}. In particular, our dataset contains analogies that are hard to solve for a non-data-driven approach (\textit{e.g.}, ``be : was :: go : went").%\todo{Added this sentence}

%%%%%%%%%%%%%%%%%%%%%%%%%%%%%%%%%%%%%%%%%%%%%%%%%
\section{Analogy Detection: Classification Model}
%(strengthening all we had in DSAA: 8 by 8 balanced data, transferebility experiments, omnilingual with more data)

%------------------------------
%\subsection{Experimental Setup}
\label{sec:clf-expe}
% expe 1: base results + significativity
% expe 2: 8 by 8
As mentioned in \cref{sec:formal-analogy}, analogy detection corresponds to a classification task, and can be used for example to mine analogical grids \cite{tools-analogy:2018:fam-lepage} which can help in language understanding and learning as well as rule extraction by inference.
Using the training procedure described in \cref{sec:training} we train an analogy classifier for each of the 11 languages in the the datasets mentioned above.
The results are reported in \cref{tab:clf-results} and detailed below.

\paragraph{Baselines.}We report the results of three state of the art methods described in \cref{sec:formal-analogy} as baselines: the classifier from \cite{tools-analogy:2018:fam-lepage} and the two regression models of \cite{analogy-alea:2009:langlais} and \cite{minimal-complexity:2020:murena}, that we respectively refer to as \textit{Lepage}, \textit{alea} and \textit{Kolmo}.
To classify a given analogy $A:B::C:D$ with these regression models, we generate the top $k$ (either $1$ or $10$) most likely solutions for $A:B::C:x$. If $D$ is within the solutions, we consider $A:B::C:D$ a valid analogy, otherwise we consider it invalid.
Using $k=1$ means that we only consider the best answer of the model as the expected solution. However, in some situations, multiple solutions are plausible, so taking $k=10$ offers more plausibility in that regard.
This results in two variants for the \textit{alea} and \textit{Kolmo} baselines: \textit{a@1}, \textit{a@10}, \textit{K@1}, and \textit{K@10}.
We only report the result of the best of the three baselines in \cref{tab:clf-results}, but detailed results per baseline are available in the appendix.\todo{Appendix} %In most cases, there is no significant difference between the baselines.

%----------------------
\paragraph{Results and discussion.}

\begin{table}[htb]
    \centering
        \caption{Classification accuracy (in \%) against the best performing baseline. The italicized result was obtained on 4000 analogies before data augmentation instead of 50000 as \textit{alea} and \textit{Kolmo} had much larger run time on this particular dataset.
        In this table, 
        %\textit{alea@1}, \textit{alea@10}, \textit{Kolmo@1}, \textit{Kolmo@10}, and \textit{Lepage} were respectively short-handed to \textit{a@1}, \textit{a@10}, \textit{K@1}, \textit{K@10}, and \textit{L}.
        \textit{Lepage} was short-handed to \textit{L}.
        First column is the ISO 639‑2 code of the languages, in the same order as for \cref{tab:num-analogy}.}\label{tab:clf-results}
    \begin{tabular}{l|cc|ll} 
\toprule
                  & \multicolumn{2}{c|}{\textbf{ANNc (ours)}} & \multicolumn{2}{c}{\textbf{Best Baseline}}                          \\
\textbf{}  & Valid            & Invalid              & {Valid}     & {Invalid}               \\ 
\midrule
Ar             & \textbf{99.39}  & 97.40 & 34.21           (a@10)  & \textbf{97.79}            (K@1)  \\
Fi            & \textbf{99.58}  & 97.98 & 25.60           (L)   & \textit{\textbf{98.78}}   (a@1)   \\
Ka           & \textbf{99.87}  & 92.06 & 93.20           (K@10) & \textbf{95.21}            (a@1)   \\
De             & \textbf{99.42}  & 95.24 & 86.90           (a@10)  & \textbf{97.19}                     (a@1)   \\
Hu          & \textbf{99.84}  & \textbf{98.71} & 36.80           (K@10) & 98.40                     (K@1)  \\
% Mt            & \textbf{100.00} & \textbf{98.61} & 78.05           (a@10)  & 69.29                    (K@1)  \\
% Nv             & \textbf{99.88}  & 77.79 & 21.45           (K@10) & \textbf{94.93}            (K@1)  \\
% Ru            & \textbf{99.00}  & 93.50 & 42.37           (a@10)  & \textbf{93.88}           (L)   \\
% Es            & \textbf{96.89}  & \textbf{89.96} & 85.90           (a@10)  & {86.62}         (L)   \\
% Tr            & \textbf{99.69}  & 82.03 & 44.76           (a@10)  & \textbf{91.40}           (K@1)  \\ 
% \midrule
% Jp           & \textbf{99.70}  & 91.93 & 19.20           (K@10) & \textbf{98.13}            (L)   \\
Mt            & \textbf{99.88}  & \textbf{77.79} & 78.05           (a@10)  & 69.29                    (K@1)  \\
Nv             & \textbf{99.00}  & 93.50 & 21.45           (K@10) & \textbf{94.93}            (K@1)  \\
Ru            & \textbf{96.89}  & {89.96} & 42.37           (a@10)  & \textbf{93.88}           (L)   \\
Es            & \textbf{99.69}  & 82.03 & 85.90           (a@10)  & \textbf{86.62}         (L)   \\
Tr            & \textbf{99.70}  & \textbf{91.93} & 44.76           (a@10)  & {91.40}           (K@1)  \\ 
\midrule
Jp           & \textbf{100.00} & \textbf{98.61} & 19.20           (K@10) & {98.13}            (L)   \\
\bottomrule
\end{tabular}
    \end{table}
    
%p-value between CNN_Valid and Baseline_Valid < 0.01 (0.000266440519100945)
%p-value between CNN_Invalid and Baseline_Invalid > 0.1 (0.868869734173962)
Our ANNc model significantly outperforms the best baselines for valid analogies (paired Student t-test $p\approx 0.00027$) and there is no significant difference for invalid analogies (paired Student t-test $p\approx 0.869$).
For each language, we observe a p-value $p<10^{-10}$ for the 1-sampled t-test against the baselines.%\todo{add to appendix if we manage to have them or remove if we do not have the results}.

We notice that the classifier \textit{Lepage} is the best in only $4$ of $22$ cases. This result was not expected, since the method has been developed as a classifier, while \textit{alea} and \textit{Kolmo} are adapted regression models. 

%While it would make sense if the best baseline was the classifier \textit{Lepage} and not the adapted regression models \textit{alea} and \textit{Kolmo}, \textit{Lepage} is the best in only $4$ of $22$ cases.

Our model manages to capture the features of morphological analogy using an embedding model adapted for morphological tasks and a classifier.
Results on transferability experiments in~\cite{transfer:2021:alsaidi} also indicate that the process of analogy itself is correctly embedded in the classifier part, while the morphological information is encoded almost exclusively in the embedding part of the model.

%%%%%%%%%%%%%%%%%%%%%%%%%%%%%%%%%%%%%%%%%%%
\section{Analogy Solving: Regression Model}
%(all the new stuff)

%------------------------------
%\subsection{Experimental Setup}
\label{sec:reg-expe}
Analogy solving paves way to understanding cognitive processes~\cite{analogy-making-cognition:2001:mitchel} and allows the generation of new but coherent and understandable words like \textit{``bestest''}.
We use the training procedure described in \cref{sec:training} and the same datasets as for classification to train analogy regression models. We train 3 models with different random initialization for each of the 11 languages available.
Note that we use only valid analogies to train the regression models.
The results are reported in \cref{tab:reg-results} and discussed further in this section.

\paragraph{Baselines.}
We use the regression models \textit{alea}~\cite{analogy-alea:2009:langlais} and \textit{Kolmo}~\cite{minimal-complexity:2020:murena} mentioned in \cref{sec:clf-expe} as baselines.
\footnote{The regression accuracy reported in \cref{tab:reg-results} is the same as the classification results of \textit{alea@1} and \textit{Kolmo@1} for valid analogies. Indeed, the procedure to compute the regression accuracy is the same as the one to obtain a classification with $k=1$.}

%----------------------
\paragraph{Results and discussion.}
\begin{table}[h]
    \centering
    \caption{Accuracy (in \%) of 3 runs of the regression model against the best baseline (Student t-test: ***~$p< 0.01$; **~$p\in ]0.01,0.05]$; *~$p\in ]0.05,0.1]$).}
    \label{tab:reg-results}
    \begin{tabular}{lllc}
    \toprule
    & \textbf{ANNr (ours)} & & \\
    \textbf{Language} & {(mean $\pm$ std.)}& \textbf{Best baseline} &\textbf{T-test}\\
    \midrule
Arabic      & $\mathbf{77.97} \pm 16.03$ & $28.94$ (Kolmo) & **\\
Finnish     & $\mathbf{37.78} \pm 9.28$ & $22.82$ (Kolmo) & *\\
Georgian    & $\mathbf{94.66} \pm 1.13$ & $86.81$ (alea) & \\
German      & $\mathbf{86.38} \pm 0.45$ & $85.46$ (alea) & *\\
Hungarian   & $\mathbf{53.83} \pm 3.12$ & $35.79$ (alea) & **\\
Maltese     & $\mathbf{75.00} \pm 5.08$ & $74.49$ (alea) & \\
Navajo      &$\mathbf{31.74} \pm 0.90$ & $17.97$ (Kolmo) & ***\\
Russian     & $\mathbf{75.15} \pm 0.44$ & $42.00$ (alea) & ***\\
Spanish     & $\mathbf{86.27} \pm 0.71$ & $85.23$ (alea) & \\
Turkish     & $\mathbf{61.95} \pm 10.86$ & $42.53$ (alea) & *\\
\midrule
Japanese    & $\mathbf{61.60} \pm 1.33$ & $18.62$ (Kolmo) & ***\\
    \bottomrule
    \end{tabular}

\end{table} 

If in some cases there is no significant difference between the best baseline and our regression model (Student t-test $p>0.1$), our model outperforms the baseline with a more or less significant difference. In some cases, the gap is larger than a significant 49\% ($p<0.01$).

While these results are encouraging, the results presented in \cref{sec:dropout} and some of our latest unpublished experiments show a significant margin of improvement of the model through simple means, like larger embedding models, or a full autoencoder architecture for the embedding model.
Indeed, while our regression model is unable to generate new words from embeddings yet, further work on the embedding model to equip it with a decoder could allow us to directly generate new words. More importantly in terms of performance, this would allow us to add to the criterion a loss term corresponding to the generated word.

The generation space of our regression model can be seen as closed since it can only generate words from the vocabulary.
This contrasts with the more open generation spaces of \textit{Kolmo} and \textit{alea}, which are able to generate new words.
This difference might be one factor of the significant gap in performance we observe, and further experiments using a decoder would allow us to determine how important this factor is in the performance.

%%%%%%%%%%%%%%%%%%%%%%%%%%%%%%%%%%%%%%
\section{Sensibility to Input Perturbations}\label{sec:dropout}
\begin{figure*}[htb]
    \centering
    \includegraphics[width=.9412\textwidth]{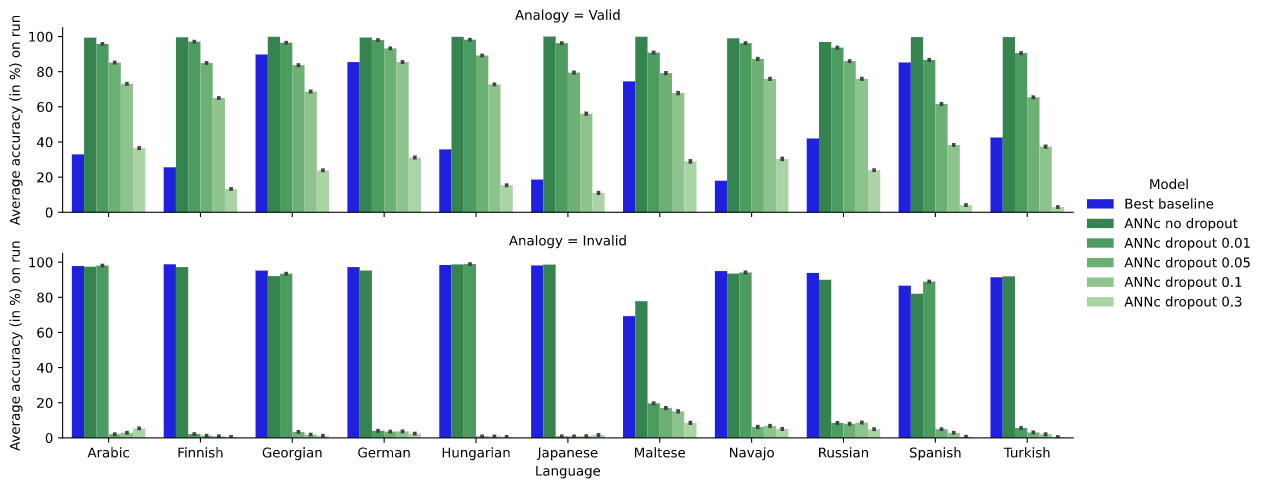}
    \caption{Classification accuracy (in \%) for each language with input perturbation using various dropout probabilities. Error bars correspond to standard deviation.}
    \label{fig:clf-dropout}
\end{figure*}
\begin{figure*}[htb]
    \centering
    \includegraphics[width=.9412\textwidth]{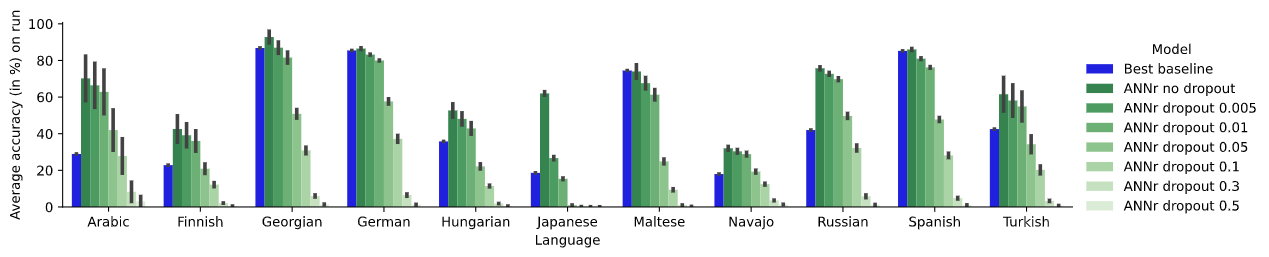}
    \caption{Regression accuracy (in \%) for each language with input perturbation using various dropout thresholds. For each of the 3 models of each language, the results with each dropout probability is significantly different from the other probabilities (Student t-test $p < 0.01$). Error bars correspond to standard deviation.}
    \label{fig:reg-dropout}
\end{figure*}

To evaluate the epistemic uncertainty~\cite{uncertainty:2016:gal} of both the ANNc and ANNr, we test their sensibility to perturbations of the input, by introducing random dropout on the embeddings during evaluation.

%------------------------------
%\subsection{Experimental Setup}
\paragraph{Input perturbation with dropout.}
For classification and regression, we introduce a random dropout\footnote{Applying a random (zero) dropout with a probability $p_d$ on a vector means that each component of the vector has a probability $p_d$ to be replaced by $0$.} between the embedding model and, respectively, the classification and regression models.
This dropout serves as a perturbation of the input, as meaningful information contained in the embedding is replaced by 0.
We evaluate the performance of the models with several dropout probabilities: $0.01$, $0.05$, $0.1$, $0.3$ for classification, as well as $0.005$ and $0.5$ for regression. For each dropout value, the experiment is repeated 10 times to determine the impact of the randomness of dropout. For the 3 regression models trained, we repeat the experiment 10 times per model.

\paragraph{Results and discussion.}
The results are summarized in \cref{fig:clf-dropout,fig:reg-dropout}, and we provide more detailed results in appendix\todo{appendix here - Added sign. test for classif with dropout}.
For each regression and classification model, the results with each dropout probability is significantly different (Student t-test $p<0.01$) from adjacent probabilities (\textit{e.g.}, with regression model 1 on Arabic, the performance without dropout is significantly different than with $p_d=0.005$, which is significantly different than with $p_d=0.01$). We note that the random aspect of the dropout process does not bring much variance in the results, as it can be seen with the error bars of \cref{fig:clf-dropout} that represent the standard deviation. In \cref{fig:reg-dropout}, the error bars are almost exclusively due to inter-model variance between the 3 models trained.

For both the regression and classification models, we observe that the more perturbed the input is, the lower the performance.
This kind of behavior is what we would expect from models relying on their input to produce the output.
Additionally, an effect particularly visible for the classification of invalid analogies is that there is a large drop in performance past a certain threshold, but that threshold is not the same for all models.
For example, for Arabic and Hungarian the threshold is between $p_d=0.05$ and $p_d=0.1$, while for Maltese and German it is between $p_d=0.01$ and $p_d=0.05$. This effect is also present in the regression and the classification of valid analogies, even if harder to notice.
This behavior can be interpreted as the model managing to perform well even when $1\%$ of the embedding is missing, with the example of invalid analogies in Arabic.
That means that those $1\%$ most likely contain redundant information. However, once we remove $5\%$, the model performs much poorly, so these $5\%$ contain necessary information. In practical terms, this indicates that we could most likely reduce the embedding size by $1\%$ without affecting the classification performance much.
For regression, some languages, and in particular Japanese, do not show this characteristic ``step'', thus, by extrapolation, we could expect the performance to increase with larger embedding models.
%-------------------
%\subsection{Results}

% --------------
\section{Conclusion and Perspectives}
%While our model does not offer the same level of interpretability as symbolic methods like the one by Murena~\textit{et~al.}~\cite{minimal-complexity:2020:murena}, which provides explicit morphological transformations as an additional result, we obtain significantly higher performances on analogy detection and solving.

Symbolic methods like the one by Murena~\textit{et~al.}~\cite{minimal-complexity:2020:murena} provide explicit morphological transformations as an additional result. While our model does not offer the same level of interpretability as those symbolic approaches, we obtain significantly higher performances on analogy detection and solving with a simple and shallow deep learning model.

As mentioned in \cref{sec:reg-expe,sec:dropout}, there are multiple plausible improvements to our approach, in particular for the regression model. Indeed, to our best knowledge, there exists no decoder for most word embedding models, and our morphological embedding model among them. Building a model able to generate coherent and potentially new words from a latent representation would improve our regression model.

Furthermore, our embedding model offers insights in the morphology of the languages we manipulate, as it captures morphological information to solve analogies.
To expand our understanding, we will perform a qualitative analysis of the patterns captured by the CNN filters.

A qualitative analysis of the ANNc and ANNr model parameters would shed light on which numerical relations are used when dealing with analogy in the latent space, thus allowing to complement the simple arithmetic operations used traditionally~\cite{embedding-analogies:2016:drozd}.

%Finally, extending the model to applications in other domains is a key task, albeit challenging, to model analogy.
Finally, applications in other domains is a key task, albeit challenging, to model analogy.
An example of such applications would be dataset augmentation, as mentioned in \cite{ap-functions-boolean-examples:2017:couceiro}.
The results we obtain, along with those of~\cite{analogies-ml-perspective:2019:lim,transfer:2021:alsaidi}, support our intuition that the simple deep learning framework discussed in this paper is not specific to analogies on words, and support the plausibility of extending our model to applications in other domains.

\section*{Acknowledgments}

This research was partially supported by TAILOR, a project funded by EU Horizon 2020 research and innovation programme under GA No 952215, and the Inria Project Lab ``Hybrid Approaches for Interpretable AI'' (HyAIAI).

The code used for our experiments is written in Python 3.9 and PyTorch.
Experiments presented in this paper were carried out using computational clusters equipped with GPU from the Grid'5000 testbed (see \url{https://www.grid5000.fr}). Namely, the \textit{grele} and \textit{grue} clusters were used.

\bibliographystyle{unsrtnat}
 \bibliography{bib.bib}

\begin{thebibliography}{28}
\providecommand{\natexlab}[1]{#1}
\providecommand{\url}[1]{\texttt{#1}}
\expandafter\ifx\csname urlstyle\endcsname\relax
  \providecommand{\doi}[1]{doi: #1}\else
  \providecommand{\doi}{doi: \begingroup \urlstyle{rm}\Url}\fi

\bibitem[Pennington et~al.(2014)Pennington, Socher, and
  Manning]{glove:2014:pennington}
Jeffrey Pennington, Richard Socher, and Christopher~D. Manning.
\newblock Glove: Global vectors for word representation.
\newblock In Alessandro Moschitti, Bo~Pang, and Walter Daelemans, editors,
  \emph{EMNLP}, pages 1532--1543, 2014.

\bibitem[Mikolov et~al.(2013)Mikolov, Chen, Corrado, and
  Dean]{efficient-representation-w2v:2013:mikolov}
Tom{\'{a}}s Mikolov, Kai Chen, Greg Corrado, and Jeffrey Dean.
\newblock Efficient estimation of word representations in vector space.
\newblock In Yoshua Bengio and Yann LeCun, editors, \emph{1st ICLR, Workshop
  Track}, Scottsdale, Arizona, USA, 2013.

\bibitem[Bojanowski et~al.(2016)Bojanowski, Grave, Joulin, and
  Mikolov]{fasttext:2016:bojanwoski}
Piotr Bojanowski, Edouard Grave, Armand Joulin, and Tom{\'{a}}s Mikolov.
\newblock Enriching word vectors with subword information.
\newblock \emph{CoRR}, abs/1607.04606, 2016.
\newblock URL \url{http://arxiv.org/abs/1607.04606}.

\bibitem[Drozd et~al.(2016)Drozd, Gladkova, and
  Matsuoka]{embedding-analogies:2016:drozd}
Aleksandr Drozd, Anna Gladkova, and Satoshi Matsuoka.
\newblock Word embeddings, analogies, and machine learning: Beyond king - man +
  woman = queen.
\newblock In \emph{26th {COLING}}, pages 3519--3530, December 2016.

\bibitem[Chen et~al.(2017)Chen, Peterson, and
  Griffiths]{eval-vector-analogy:2017:chen}
Dawn Chen, Joshua~C. Peterson, and Tom Griffiths.
\newblock Evaluating vector-space models of analogy.
\newblock In Glenn Gunzelmann, Andrew Howes, Thora Tenbrink, and Eddy~J.
  Davelaar, editors, \emph{39th CogSci}, London, UK, 2017. Cognitive Science
  Society.

\bibitem[Prade and Richard(2014)]{trends-analogical-reasoning:2014:prade}
Henri Prade and Gilles Richard.
\newblock A short introduction to computational trends in analogical reasoning.
\newblock In Henri Prade and Gilles Richard, editors, \emph{Computational
  Approaches to Analogical Reasoning: Current Trends}, volume 548 of
  \emph{Studies in Computational Intelligence}, pages 1--22. Springer, 2014.

\bibitem[Miclet et~al.(2008)Miclet, Bayoudh, and
  Delhay]{analogical-dissimilarity:2008:miclet}
Laurent Miclet, Sabri Bayoudh, and Arnaud Delhay.
\newblock Analogical dissimilarity: Definition, algorithms and two experiments
  in machine learning.
\newblock \emph{Journal Artificial Intelligence Research}, 32:\penalty0
  793--824, 2008.
\newblock \doi{10.1613/jair.2519}.

\bibitem[Lim et~al.(2019)Lim, Prade, and
  Richard]{analogies-ml-perspective:2019:lim}
Suryani Lim, Henri Prade, and Gilles Richard.
\newblock Solving word analogies: {A} machine learning perspective.
\newblock In Gabriele Kern{-}Isberner and Zoran Ognjanovic, editors, \emph{15th
  ECSQARU}, volume 11726, pages 238--250, 2019.

\bibitem[Langlais et~al.(2009)Langlais, Yvon, and
  Zweigenbaum]{analogy-alea:2009:langlais}
Philippe Langlais, Fran{\c{c}}ois Yvon, and Pierre Zweigenbaum.
\newblock Improvements in analogical learning: Application to translating
  multi-terms of the medical domain.
\newblock In Alex Lascarides, Claire Gardent, and Joakim Nivre, editors,
  \emph{12th {EACL}}, pages 487--495, Athens, Greece, 2009. ACL.

\bibitem[Fam and Lepage(2018)]{tools-analogy:2018:fam-lepage}
Rashel Fam and Yves Lepage.
\newblock Tools for the production of analogical grids and a resource of n-gram
  analogical grids in 11 languages.
\newblock In Nicoletta Calzolari, Khalid Choukri, Christopher Cieri, Thierry
  Declerck, Sara Goggi, K{\^{o}}iti Hasida, Hitoshi Isahara, Bente Maegaard,
  Joseph Mariani, H{\'{e}}l{\`{e}}ne Mazo, Asunci{\'{o}}n Moreno, Jan Odijk,
  Stelios Piperidis, and Takenobu Tokunaga, editors, \emph{11th {LREC}},
  Miyazaki, Japan, 2018. ELRA.

\bibitem[Murena et~al.(2020)Murena, Al-Ghossein, Dessalles, and
  Cornuéjols]{minimal-complexity:2020:murena}
Pierre-Alexandre Murena, Marie Al-Ghossein, Jean-Louis Dessalles, and Antoine
  Cornuéjols.
\newblock Solving analogies on words based on minimal complexity
  transformation.
\newblock In Christian Bessiere, editor, \emph{29th IJCAI}, pages 1848--1854,
  2020.

\bibitem[de~Saussure(1916)]{linguistique-generale:1916:saussure}
Ferdinand de~Saussure.
\newblock \emph{Cours de linguistique g{\'e}n{\'e}rale}.
\newblock Payot, Paris, 1916.

\bibitem[Lepage and Ando(1996)]{lepage1996saussurian}
Yves Lepage and Shinichi Ando.
\newblock Saussurian analogy: a theoretical account and its application.
\newblock In \emph{16th COLING}, 1996.

\bibitem[Lepage(2003)]{analogy-commutation-linguistic-fr:2003:lepage}
Yves Lepage.
\newblock \emph{{De l'analogie rendant compte de la commutation en
  linguistique}}.
\newblock Habilitation {\`a} diriger des recherches, {Universit{\'e}
  Joseph-Fourier - Grenoble I}, May 2003.

\bibitem[Bayoudh et~al.(2012)Bayoudh, Prade, and
  Richard]{bayoudh2012evaluation}
Meriam Bayoudh, Henri Prade, and Gilles Richard.
\newblock Evaluation of analogical proportions through kolmogorov complexity.
\newblock \emph{Knowledge-Based Systems}, 29:\penalty0 20--30, 2012.

\bibitem[Fam and Lepage(2016)]{fam2016morphological}
Rashel Fam and Yves Lepage.
\newblock Morphological predictability of unseen words using computational
  analogy.
\newblock In \emph{ICCBR Workshops}, pages 51--60, 2016.

\bibitem[Lepage(2017)]{sigmorphon:2017:lepage}
Yves Lepage.
\newblock Character-position arithmetic for analogy questions between word
  forms.
\newblock In Antonio~A. S{\'{a}}nchez{-}Ruiz and Anders Kofod{-}Petersen,
  editors, \emph{25th ICCBR (workshops)}, volume 2028, pages 23--32, 2017.

\bibitem[Yvon(2003)]{finite-state-trancducers:2003:yvon}
Fran{\c{c}}ois Yvon.
\newblock Finite-state transducers solving analogies on words.
\newblock \emph{Rapport GET/ENST\&LTCI}, 2003.

\bibitem[Rumelhart and Abrahamson(1973)]{rumelhart1973model}
David~E Rumelhart and Adele~A Abrahamson.
\newblock A model for analogical reasoning.
\newblock \emph{Cognitive Psychology}, 5\penalty0 (1):\penalty0 1--28, 1973.

\bibitem[Cotterell et~al.(2016)Cotterell, Kirov, Sylak-Glassman, Yarowsky,
  Eisner, and Hulden]{sigmorphon:2016:cotterell}
Ryan Cotterell, Christo Kirov, John Sylak-Glassman, David Yarowsky, Jason
  Eisner, and Mans Hulden.
\newblock The sigmorphon 2016 shared task---morphological reinflection.
\newblock In Micha Elsner and Sandra K{\"{u}}bler, editors, \emph{SIGMORPHON,
  2016}, Berlin, Germany, August 2016. ACL.

\bibitem[Vania(2020)]{morpho-thesis:2020:vania}
Clara Vania.
\newblock \emph{On understanding character-level models for representing
  morphology}.
\newblock PhD thesis, University of Edinburgh, 2020.

\bibitem[Kingma and Ba(2015)]{adam:2014:kingma}
Diederik~P. Kingma and Jimmy Ba.
\newblock Adam: {A} method for stochastic optimization.
\newblock In Yoshua Bengio and Yann LeCun, editors, \emph{3rd {ICLR}}, 2015.

\bibitem[Alsaidi et~al.(2021{\natexlab{a}})Alsaidi, Decker, Lay, Marquer,
  Murena, and Couceiro]{transfer:2021:alsaidi}
Safa Alsaidi, Amandine Decker, Puthineath Lay, Esteban Marquer,
  Pierre-Alexandre Murena, and Miguel Couceiro.
\newblock {On the Transferability of Neural Models of Morphological Analogies}.
\newblock In \emph{{AIMLAI, ECML PKDD 2021: European Conference on Machine
  Learning and Principles and Practice of Knowledge Discovery in Databases}},
  Bilbao/Virtual, Spain, September 2021{\natexlab{a}}.
\newblock URL \url{https://hal.inria.fr/hal-03313591}.

\bibitem[Karpinska et~al.(2018)Karpinska, Li, Rogers, and
  Drozd]{jap-data:2018:karpinska}
Marzena Karpinska, Bofang Li, Anna Rogers, and Aleksandr Drozd.
\newblock Subcharacter information in japanese embeddings: when is it worth it?
\newblock In \emph{Workshop on the Relevance of Linguistic Structure in Neural
  Architectures for NLP}, pages 28--37, Melbourne, Australia, 2018. ACL.

\bibitem[Alsaidi et~al.(2021{\natexlab{b}})Alsaidi, Decker, Lay, Marquer,
  Murena, and Couceiro]{alsaidi:hal-03313556}
Safa Alsaidi, Amandine Decker, Puthineath Lay, Esteban Marquer,
  Pierre-Alexandre Murena, and Miguel Couceiro.
\newblock {A Neural Approach for Detecting Morphological Analogies}.
\newblock In \emph{{The 8th IEEE International Conference on Data Science and
  Advanced Analytics (DSAA)}}, Porto/Online, Portugal, October
  2021{\natexlab{b}}.
\newblock URL \url{https://hal.inria.fr/hal-03313556}.

\bibitem[Mitchell(2001)]{analogy-making-cognition:2001:mitchel}
Melanie Mitchell.
\newblock Analogy making as a complex adaptive system.
\newblock In \emph{Santa Fe Institute Studies in the Sciences of Complexity},
  pages 335--360. Reading, Mass.; Addison-Wesley; 1998, 2001.

\bibitem[Gal(2016)]{uncertainty:2016:gal}
Yarin Gal.
\newblock \emph{Uncertainty in Deep Learning}.
\newblock PhD thesis, University of Cambridge, 2016.

\bibitem[Couceiro et~al.(2017)Couceiro, Hug, Prade, and
  Richard]{ap-functions-boolean-examples:2017:couceiro}
Miguel Couceiro, Nicolas Hug, Henri Prade, and Gilles Richard.
\newblock Analogy-preserving functions: A way to extend boolean samples.
\newblock In \emph{26th {IJCAI}}, pages 1575--1581, 2017.
\newblock \doi{10.24963/ijcai.2017/218}.

\end{thebibliography}
 
\appendix
\part*{Appendices}
\section{Baselines}

    \begin{table}
        \centering
        \caption{Classification accuracy results (in \%) of the baselines. Italicized results were obtained on 4000 analogies before data augmentation instead of 50000 as \textit{alea} and \textit{Kolmo} had much larger run time on this particular dataset. Due to this same reason, results marked with -- were not computed as they required an even larger run time.}
        \label{tab:appx-clf-results}
        \begin{tabular}{l|cc|cc|cc|cc|cc}\toprule
                                & \multicolumn{2}{c|}{\textbf{Kolmo@1}} & \multicolumn{2}{c|}{\textbf{Kolmo@10}} & \multicolumn{2}{c|}{\textbf{alea@1}} & \multicolumn{2}{c|}{\textbf{alea@10}} & \multicolumn{2}{c}{\textbf{Lepage}} \\
            \textbf{Language}   &         Valid & Invalid  & Valid & Invalid& Valid & Invalid& Valid & Invalid& Valid & Invalid  \\\midrule
            Arabic      & 28.94&\textbf{97.79}&           33.55&97.68 &28.28&97.68& \textbf{34.21}&97.68    &32.94   &97.67 \\ %no significant difference for neg
            Finnish     & \textit{22.82}&\textit{98.12}&  --&-- &\textit{21.83}   &\textit{\textbf{98.78}}& --&--  &\textbf{25.60}   &98.05\\
            Georgian    & 80.19&95.01&                    \textbf{93.20}&94.61 &86.81&\textbf{95.21} &91.68&95.21 &89.78   &95.18\\ %
            German      & 55.61&96.84&                    60.27&96.65 &85.46&\textbf{97.19} &\textbf{86.90}&97.18 &83.32   &96.86\\ %
            Hungarian   & 31.21&\textbf{98.40}&                    \textbf{36.80}&98.23 &35.79&98.24  &36.58&98.24    &33.81   &98.25\\ %
            Maltese     & 68.84&\textbf{69.29}&                    73.64&67.32 &74.49&67.35  &\textbf{78.05}&67.32    &73.63   &67.32\\
            Navajo      & 17.97&\textbf{94.93}&           \textbf{21.45}&94.45 &16.16&94.76  &18.34&94.76    &17.35   &94.76\\
            Russian     & 33.37&93.66&                    36.43&93.30 &42.00&93.74  &\textbf{42.37}&93.72    &41.02   &\textbf{93.88}\\
            Spanish     & 73.86&86.59&                    81.54&86.44 &85.23&86.13  &\textbf{85.90}&86.13    &83.25   &\textbf{86.62}\\
            Turkish     & 39.37&\textbf{91.4}0&                    43.51&90.78 &42.53&90.80  &\textbf{44.76}&90.80    &34.92   &90.80\\\midrule
            Japanese    & 18.62&98.13&                    \textbf{19.20}&98.09 & 3.75&98.14  & 3.77&98.14    &3.74    &\textbf{98.16}\\
            \bottomrule
        \end{tabular}
    \end{table}

To compare the performance of our models, we use the methods presented in   \cite{minimal-complexity:2020:murena}, \cite{tools-analogy:2018:fam-lepage}, and \cite{analogy-alea:2009:langlais}.
The models of Murena~\textit{et~al.}~\cite{minimal-complexity:2020:murena} and Langlais~\textit{et~al.}~\cite{analogy-alea:2009:langlais} are designed for the problem of regression. 
We adapt them into classification methods by checking whether the regression output is equal to the expected term.
Detailed results are presented in \cref{tab:appx-clf-results}, and as can be seen, in many cases the performance of the baselines in very similar.

\subsubsection{Kolmogorov Complexity Based Approach}
We use the generation model proposed by Murena~\textit{et~al.}~\cite{minimal-complexity:2020:murena} as a first baseline.
This approach considers some transformation $f$ such that $B=f(A)$ and $f(C)$ is computable.
The transformation $f$ which is the simplest is usually the one human use to solve analogies (as explained in \cite{minimal-complexity:2020:murena}), and this ``simplest'' transformation is found by minimysing the Kolmogorov complexity of $f$.

As the minimal-complexity approach is designed as a generative model, to have a classification for a given analogy $A:B::C:D$ we generate the top $k$ most likely solutions for $A:B::C:x$, in increasing order of Kolmogorov complexity. If $D$ is within the solutions, we consider $A:B::C:D$ as a valid analogy, otherwise we consider it a non-analogy. 
We use multiple threshold values from $1$ (only the most likely answer) to $10$, resulting in multiple baselines.
As the accuracy for $1<k<10$ is strictly within the range of the accuracy for $k=1$ and $k=10$, only $k=1$ and $k=10$ are displayed in the results, that we refer to as \texttt{Kolmo@1} and \texttt{Kolmo@10}.

\subsubsection{Langlais \textit{et al.}'s alea}
We use the \textit{alea} algorithm~\cite{analogy-alea:2009:langlais} as a second baseline.
This approach performs a Monte-Carlo estimation of the solutions of an analogy, by sampling among multiple sub-transformations.
Those sub-transformations are obtained by considering the words as bags of characters, and gererating permutations of characters that are present in $B$ but not in $A$ on one side and characters of $C$ on the other.
 For each quadruple $A:B::C:D$, we generate $\rho=1000$ potential solutions for $A:B::C:x$ and select the $k$ most frequent ones.
This approach is a generation model like the approach based on Kolmogorov complexity, so we use a similar process of generating potential solution and searching the expected $D$ among them.
Similarly to with the Kolmogorov complexity based approach, we use multiple threshold values from $1$ (only the most likely answer) to $10$, and report only the results with $k=1$ and $k=10$, that we refer to as \texttt{alea@1} and \texttt{alea@10}.

\subsubsection{Fam and Lepage's Classifier}
We use the analogy classifier (\texttt{is\_analogy} in \texttt{nlg/Analogy/tests/nlg\_benchmark.py}) from Fam and Lepage's toolset \cite{tools-analogy:2018:fam-lepage} to classify analogies in the same manner as with our deep learning model.
This baseline uses a matrix based approach to align characters between words and find common sub-words, and then determine whether there is an analogy.
We refer to this baseline as \texttt{Lepage}.

%in order of mention in the article
\section{Using 3, 8 or 24 invalid analogies}
We consider 3 settings for selecting invalid analogies during training, which result in different balancing of the data:\begin{enumerate}
    \item using 24 invalid analogies for 8 valid analogies, by computing the 3 invalid analogies for each of the 8 equivalent forms of every analogy extracted from data; this results in a strong imbalance in favor of invalid analogies, with 3 times more invalid than valid analogies;
    \item using 3 invalid analogies for 8 valid analogies, by computing the 3 invalid analogies only from the base form extracted from data; this results in an imbalance in favor of valid analogies, with a bit more than twice more valid than invalid analogies;
    \item using 8 invalid analogies for 8 valid analogies, by sampling 8 of the 24 invalid analogies mentioned earlier; this results in a balancing of the data, at least in number of valid and invalid analogies.
\end{enumerate}

Our early experiments on classification followed either setting 1 (\textit{i.e.}, the same pattern as \cite{analogies-ml-perspective:2019:lim}) or setting 2, which leads to faster convergence in terms of actual training time due to using fewer samples.
Those early experiments showed an obvious impact of the training setting on the performance, as can be seen in \cref{tab:8x8-results}.
Indeed, imbalance in favor of valid analogies leads to very good results on valid analogies and poorer results on invalid analogies. Conversely, imbalance in favor of invalid analogies leads to very good results on invalid analogies and poorer results on valid analogies.
\begin{table}
    \centering
    \caption{Accuracy results (in \%) for the classification task for valid ($+$) and invalid ($-$) analogies, for 3, 24, and 8 randomly sampled invalid analogies for 8 valid analogies during training.}
    \label{tab:8x8-results}
    \begin{tabular}{l|cc|cc|cc}\toprule
                            &\multicolumn{2}{c|}{\textbf{3 invalid}} & \multicolumn{2}{c|}{\textbf{24 invalid}} & \multicolumn{2}{c}{\textbf{8 invalid}} \\
        \textbf{Language}   & $+$ & $-$ & $+$ & $-$  & $+$ & $-$ \\\midrule
        Arabic      & \textbf{99.89} & 97.52 & 95.66 & \textbf{99.39} & 99.39 & 97.40 \\
        Finnish     & {99.44} & 82.62 & 83.46 & 96.94 & \textbf{99.58} & \textbf{97.98}\\
        Georgian    & {99.83} & 91.71 & 88.89 & \textbf{99.53} & \textbf{99.87} & 92.06\\
        German      & \textbf{99.48} & 89.01 & 67.20 & 87.26 & 99.42 & \textbf{95.24}\\
        Hungarian   & \textbf{99.99} & {98.81} & 98.40 & \textbf{99.25} & 99.84 & 98.71 \\
        Maltese      & {99.53} & 90.82 & 89.29 & \textbf{97.40} & \textbf{99.88} & 77.79 \\
        Navajo     & {97.95} & 79.85 & 46.33 & \textbf{97.40} & \textbf{99.00} & 93.50\\
        Russian     & \textbf{99.94} & 78.33 &  72.66 & \textbf{99.80} & 96.89 & 89.96\\
        Spanish     & {99.48} & {92.63} & 82.95 & \textbf{99.35} & \textbf{99.69} & 82.03\\
        Turkish    & \textbf{99.99} & \textbf{98.65} &  100.00 & 98.64 & 99.70 & 91.93\\\midrule
        Japanese     & {99.96} & {77.83} & 37.91 & \textbf{98.85} & \textbf{100.00} & 98.61\\
        \bottomrule
    \end{tabular}
\end{table}

\begin{table}[!h]
    \centering
    \caption{Paired t-test on the evaluation of valid analogies. The test variable is called $t$ and the p-value $p$.}
    \label{tab:8x8-ttest-valid}
    \begin{tabular}{c|c|c}\toprule
        &8/8&8/24\\\midrule
        \multirow{2}{*}{8/3}&$t =1.782209$&$t = 3.433534$\\
        &$p = 0.105049$&$p = 0.006400$\\
        \multirow{2}{*}{8/8}&\multirow{2}{*}{/}&$t = 3.371031$\\
        &&$p = 0.007109$\\\bottomrule
    \end{tabular}
\end{table}
\begin{table}[!h]
    \centering
    \caption{Paired t-test on the evaluation of invalid analogies. The test variable is called $t$ and the p-value $p$.}
    \label{tab:8x8-ttest-invalid}
    \begin{tabular}{c|c|c}\toprule
        &8/8&8/24\\\midrule
        \multirow{2}{*}{8/3}&
        $t = -45.802195$&
        $t = -22.038142$\\
        &$p = 5.926135\times10^{-13}$
        &$p = 8.294267\times10^{-10}$\\
        \multirow{2}{*}{8/8}&\multirow{2}{*}{/}&
        $t = -2.148662$\\
        &&$p = 0.057193$\\\bottomrule
    \end{tabular}
\end{table}

This led us to the choice of randomly sampling 8 invalid analogies from the 24 available, which restored balance while keeping as much variety of invalid analogies as possible.
On valid analogies, this new setting (setting 3 in the list above) proved to produce results significantly better than in setting 1 ($p<0.01$), and there was no significant difference from setting 2 ($p>0.1$).
On invalid analogies, setting 3 was significantly better than setting 2. ($p<0.01$).
Finally, there is a slight difference on invalid analogies between settings 2 and 3 ($0.05<p<0.1$), even if the direction differs depending on the language.
Detailed paired Student t-tests are reported in 
\cref{tab:8x8-ttest-valid,tab:8x8-ttest-invalid}.
Overall, this new setting 3 with balanced invalid and valid analogies showed comparable results with the best of settings 1 and 2 on both valid and invalid analogies.

% \begin{table}[htbp]
%     \centering
%     \begin{tabular}{c|c|c}\toprule
%         &8/8&8/24\\\midrule
%         \multirow{2}{*}{8/3}&t =1.782209020740016&t = 3.433533544118359\\
%         &p-value = 0.10504931312151063&p-value = 0.006400438973654351\\
%         \multirow{2}{*}{8/8}&\multirow{2}{*}{/}&t = 3.3710313093051627\\
%         &&p-value = 0.007109075954128679\\
%         \multirow{2}{*}{8/24}&\multirow{2}{*}{/}&\multirow{2}{*}{/}\\
%         &&\\\bottomrule
%     \end{tabular}
%     \caption{Paired t-test on the evaluation of valid analogies}
%     \label{tab:8x8-ttest-valid}
% \end{table}

% \begin{table}
%     \centering
%     \begin{tabular}{c|c|c}\toprule
%         &8/8&8/24\\\midrule
%         \multirow{2}{*}{8/3}&t = -45.80219462459328&t = -22.038141847976163\\
%         &p-value = 5.926135488111583e-13&p-value = 8.294266626953892e-10\\
%         \multirow{2}{*}{8/8}&\multirow{2}{*}{/}&t = -2.1486623541590513\\
%         &&p-value = 0.05719337920523426\\
%         \multirow{2}{*}{8/24}&\multirow{2}{*}{/}&\multirow{2}{*}{/}\\
%         &&\\\bottomrule
%     \end{tabular}
%     \caption{Paired t-test on the evaluation of invalid analogies}
%     \label{tab:8x8-ttest-invalid}
% \end{table}

\section{Performance on Classification with Dropout}

In \cref{tab:ind_ttest_valid,tab:ind_ttest_invalid} we provide statistical tests on the significance of the difference of the results displayed in Figure 2 of the main paper.

\begin{table}[!h]
    \centering
    \begin{tabular}{c|ccc}\toprule
        &0.05&0.1& 0.3\\\midrule
        \multirow{1}{*}{0.01}& $0.00045$ & $5.00110\times10^{-6}$ & $6.44357\times10^{-15}$\\
        \multirow{1}{*}{0.05}&\multirow{1}{*}{/}&$0.00765$ & $1.56664\times10^{-11}$\\
        \multirow{1}{*}{0.1}&\multirow{1}{*}{/}&\multirow{1}{*}{/} & $1.74502\times10^{-7}$\\\bottomrule
    \end{tabular}
    \caption{Independent t-test on the evaluation of valid analogies with different dropout settings.}
    \label{tab:ind_ttest_valid}
\end{table}

\begin{table}[!h]
    \centering
    \begin{tabular}{c|ccc}\toprule
        &0.05&0.1& 0.3\\\midrule
        \multirow{1}{*}{0.01}& $0.00701$ & $0.00653$ & $0.00512$\\
        \multirow{1}{*}{0.05}&\multirow{1}{*}{/}& $0.83337$ & $0.28144$\\
        \multirow{1}{*}{0.1}&\multirow{1}{*}{/}&\multirow{1}{*}{/} & $0.38175$\\\bottomrule
    \end{tabular}
    \caption{Independent t-test on the evaluation of invalid analogies with different dropout settings.}
    \label{tab:ind_ttest_invalid}
\end{table}

\end{document}